\documentclass{article}

\usepackage[final]{neurips_2021}

\usepackage[utf8]{inputenc} 
\usepackage[T1]{fontenc}    
\usepackage{hyperref}       
\usepackage{url}            
\usepackage{booktabs}       
\usepackage{amsfonts}       
\usepackage{nicefrac}       
\usepackage{microtype}      
\usepackage{xcolor}         

\usepackage{graphicx}
\usepackage{amsmath}
\usepackage{amssymb}
\usepackage{amsbsy}
\usepackage{textcomp}
\usepackage{bbm}
\usepackage{bm}
\usepackage{multirow}
\usepackage{tabularx}
\usepackage{ragged2e}
\usepackage{booktabs}
\usepackage{subfig}
\usepackage{tablefootnote}
\usepackage{footnotehyper}
\usepackage{xspace}
\usepackage{enumitem}
\usepackage{multirow}
\usepackage{wrapfig}

\newlength{\oldintextsep}
\setcitestyle{square,numbers}
\hypersetup{colorlinks,linkcolor={blue},citecolor={blue},urlcolor={purple}}

\newcolumntype{P}[1]{>{\centering\arraybackslash}p{#1}}

\usepackage{stmaryrd}

\newcommand{\howtohundredM}[1]{HowTo100M}
\newcommand{\wikihundredthree}[1]{Wiki103}

\newcommand{\methodname}[1]{\textsc{VidLanKD}}
\newcommand{\berttwelve}[1]{BERT$_{\text{12L/768H}}$}
\newcommand{\bertsix}[1]{BERT$_{\text{6L/512H}}$}
\newcommand{\LMteacher}[1]{$\operatorname{LM}^{T}$}
\newcommand{\LMstudent}[1]{$\operatorname{LM}^{S}$}
\newcommand{\Visteacher}[1]{$\operatorname{V}^{T}$}
\newcommand{\frameencoder}[1]{$\operatorname{E}$}

\newcommand{\teachermodel}[1]{$\operatorname{M}^{T}$}
\newcommand{\studentmodel}[1]{$\operatorname{M}^{S}$}

\newcommand{\vldataset}[1]{$D_{\operatorname{VL}}$}
\newcommand{\ldataset}[1]{$D_{\operatorname{L}}$}
\newcommand{\sentence}[1]{$x$}
\newcommand{\video}[1]{$v$}

\title{\methodname{}: Improving Language Understanding via Video-Distilled Knowledge Transfer}

\author{Zineng Tang \quad Jaemin Cho \quad Hao Tan \quad Mohit Bansal
\\ 
  UNC Chapel Hill\\ 
  \texttt{\{terran, jmincho, haotan, mbansal\}@cs.unc.edu}
}

\begin{document}

\maketitle

\begin{abstract}
Since visual perception can give rich information beyond text descriptions for world understanding, there has been increasing interest in leveraging visual grounding for language learning.
Recently, vokenization \cite{tan2020vokenization} has attracted attention by
using the predictions of a text-to-image retrieval
model as labels for language model supervision.
Despite its success, the method suffers from approximation error of using finite image labels and the lack of vocabulary diversity of a small image-text dataset.
To overcome these limitations,
we present \methodname{}, a video-language knowledge distillation method for improving language understanding.
We train a multi-modal teacher model on a video-text dataset,
and then transfer its knowledge to a student language model with a text dataset.
To avoid approximation error, we propose to use different knowledge distillation objectives.
In addition, the use of a large-scale video-text dataset helps learn diverse and richer vocabularies.
In our experiments, \methodname{} achieves consistent improvements over text-only language models and vokenization models, on several downstream language understanding tasks including GLUE, SQuAD, and SWAG.
We also demonstrate the improved world knowledge, physical reasoning, and temporal reasoning capabilities of our model by evaluating on the GLUE-diagnostics, PIQA, and TRACIE datasets.
Lastly, we present comprehensive ablation studies as well as visualizations of the learned text-to-video grounding results of our teacher and student language models.\footnote{Code and models: \url{https://github.com/zinengtang/VidLanKD}}

\end{abstract}
\section{Introduction}
\label{sec:intro}
Language learning can be aided by grounded visual cues,
as they provide powerful signals for modeling a vastness of experiences in the world that cannot be documented by text alone \cite{bisk2020experience, HARNAD1990335, bender-koller-2020-climbing}.
While the recent trend of large-scale language model pretraining indirectly provides some world knowledge from text, most large text corpora (e.g., Wikipedia) do not provide enough multi-modal grounding information.
Previous works have explored multiple ways of grounding language to visual information
such as constructing a common vector space~\cite{kiela2017learning, bordes2020incorporating} and supervising the model with token-wise generated vision labels \cite{tan2020vokenization}.
However, the widely-used image-text datasets (e.g., MS COCO \cite{lin2014microsoft}) are much smaller than text-only corpora in terms of word counts
and vocabulary diversity for language learning.

The recent method of `vokenization' \cite{tan2020vokenization} is a 
promising initial
step towards addressing this problem by supervising language models with weakly-aligned vision-language groundings.
Firstly, an image-text matching model
retrieves
a corresponding image to each text token in a sentence.
Then a language model learns to predict the selected image (called `voken') for each text token.
This can be seen as a
knowledge distillation (KD) process \cite{hinton2015distilling}
from a vision-language grounding model to a language model.
Although the voken classification task helps the language model to improve on natural language understanding (NLU) tasks, there exist several limitations:
(1) images cannot faithfully convey word meanings that require more activity-based and physical commonsense knowledge. 
(2) the voken supervision suffers from approximation/quantization error
of the text-to-image retrieval.

\begin{figure*}[t]
  \centering
  \includegraphics[width=0.88\textwidth]{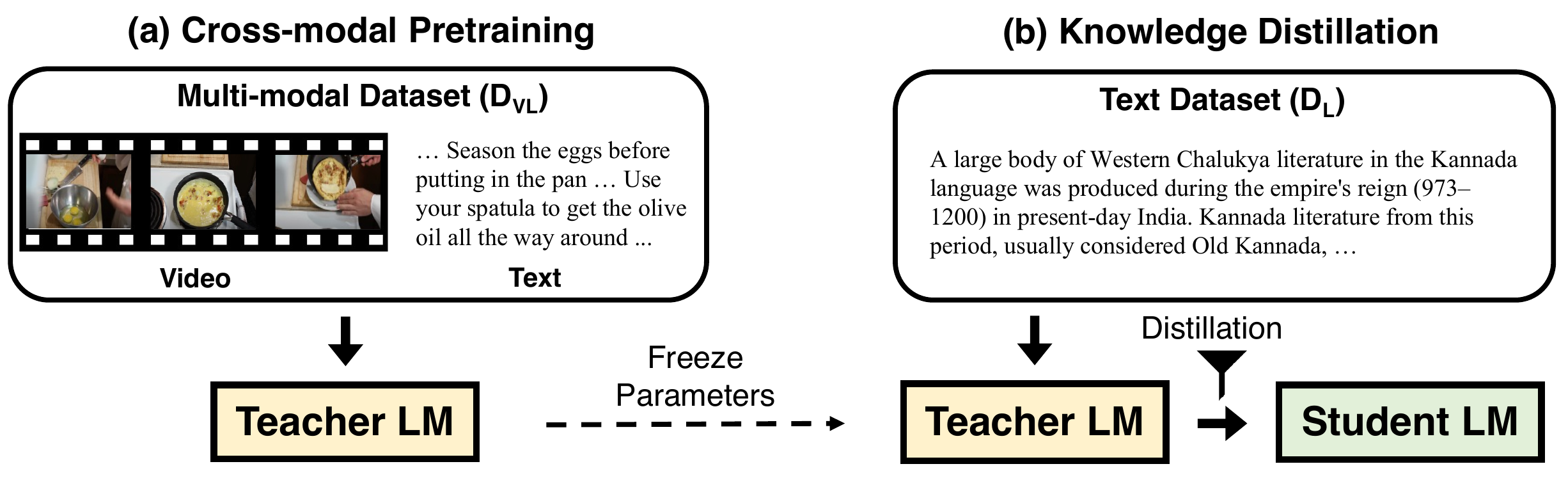}
  \vspace{-3pt}
  \caption{Overview of the proposed
    \methodname{}
  method. We first pretrain a teacher language model on a multi-modal dataset (Sec.~\ref{sec:teacher_model}). Then we distill the knowledge of the teacher model (weights frozen) to a student language model on a text dataset (Sec.~\ref{sec:student_model}).}
  \label{fig:teaser}
  \vspace{-5pt}
\end{figure*}

To address these problems,
we propose a novel Video-and-Language Knowledge Distillation method, named \methodname{}.
Our teacher model consists of a video encoder and a language encoder.
They are jointly trained with a video-language contrastive learning objective and a masked language modeling (MLM) objective on a multi-modal dataset (see Fig.~\ref{fig:teaser}).
Then, we transfer the knowledge of the frozen teacher language
encoder
to a student language model by minimizing the
distance
between contextualized text representations of two models on a text dataset.
For this, we propose to use 
different KD objectives including neuron selectivity transfer (NST)~\cite{huang2017like} and contrastive representation distillation (CRD)~\cite{tian2019contrastive}
that avoid the approximation error from voken assignments \cite{tan2020vokenization}.
For cross-modal pretraining of our teacher model, we use \howtohundredM{}~\cite{miech2019howto100m}, a large-scale video dataset which has more diverse vocabulary and
richer world commonsense (e.g., physical and temporal) knowledge
compared to MS COCO image dataset.

In our experiments, student language models learned with the proposed video-language KD objectives
outperform the baseline text-pretrained language models and the models distilled with vokenization, on several diverse natural language understanding benchmarks including GLUE \cite{wang2018glue}, SQuAD \cite{rajpurkar2016squad}, and SWAG \cite{zellers2018swag}.
We also show comprehensive ablation studies on
video encoders, student KD objectives, teacher pretraining objectives, and video vs. image-based pretraining.
Furthermore, we empirically illustrate that our model successfully learns linguistic world knowledge and physical/temporal commonsense abilities from video, by showing improved performances on the GLUE-diagnostics \cite{wang2018glue}, PIQA \cite{bisk2020piqa}, and TRACIE \cite{zhou-etal-2021-temporal} datasets.

Overall, our contributions are:
(1) a novel cross-modal knowledge distillation method for improving natural language understanding,
(2) using rich video-text data which can overcome the limitations of image vokenization,
(3) empirical improvements on several language understanding benchmarks and studying different knowledge distillation methods,
and 
(4) analysis on linguistic/physical/temporal knowledge learned from videos and ablation studies on the effectiveness of proposed components.

\section{Related Work}
\subsection{Knowledge Distillation}
Knowledge distillation (KD) \cite{hinton2015distilling} is the process of transferring knowledge from a teacher model to a student model.
It has been successfully used in a wide range of applications, such as machine translation \cite{kim2016sequence}, visual recognition \cite{he2019bag}, speech recognition \cite{chebotar2016distilling}, and recommendation systems \cite{tang2018ranking}.
Recent works advanced the field of knowledge distillation by proposing new architectures \cite{xu2017training, zhang2018deep, anil2018large, mirzadeh2020improved} and objectives \cite{huang2017like, chen2018darkrank}.

While many KD works study the problem of knowledge transfer within the same modality,
cross-modal knowledge distillation \cite{gupta2016cross, do2019compact, tian2019contrastive} tackles the knowledge transfer across different modalities. \citet{gupta2016cross} transfers the knowledge of a model trained with RGB images to another model for depth maps and optical flow.
\citet{do2019compact} proposes a KD method for visual question answering \cite{antol2015vqa},
where the trilinear (image-question-answer) relational representation of a teacher model is transferred to
a bilinear (image-question) student model.
\citet{tian2019contrastive} combines contrastive learning and knowledge distillation to improve the knowledge transfer between different modalities.
Our \methodname{} transfers the knowledge of a multi-modal teacher model learned from a video dataset to a student language model that tackles natural language understanding tasks.

\subsection{Language Pretraining}
Large-scale pretraining of contextualized language models has seen huge success in natural language processing in recent years.
ELMo \cite{peters2018deep} proposes to pretrain and fine-tune a large recurrent language model, which improves performance on a diverse set of downstream natural language processing tasks.
BERT~\cite{devlin2018bert} improves the scalability of the pretrain/fine-tune framework by using a transformer~\cite{vaswani2017attention} language model with a masked language modeling objective.
Since then, pretraining of transformer language models has been extensively explored~\cite{liu2019roberta,yang2019xlnet,lan2019albert,dong2019unified,song2019mass,raffel2020exploring,clark2020electra}
for various natural language understanding~\cite{rajpurkar2016squad,williams2017broad,zellers2018swag,wang2018glue}
and generation tasks~\cite{graff2003english, Rush_2015, rajpurkar2016squad, see2017get}.

\subsection{Multi-modal Pretraining}
Following the success of language pretraining with transformer models, pretraining of
image-text~\cite{tan2019lxmert,lu2019vilbert,chen2019uniter,li2020unimo,zhou2020unified,li2020unicoder} and
video-text~\cite{sun2019videobert,miech2019howto100m,zhu2020actbert,miech2020end,li2020hero,tang2021decembert}
multi-modal transformers have achieved improvements on numerous multi-modal downstream tasks~\cite{antol2015vqa,xu2016msr,zhou2017towards}.
The multi-modal transformers take both visual and textual inputs and are pretrained on image-text or video-text pairs
with multi-modal masked language modeling objectives.
Despite the success on multi-modal downstream tasks, \citet{tan2020vokenization} finds that the multi-modal pretraining does not improve (and sometimes even harms) the language understanding performance. This is because the scale and diversity of text vocabulary of image-text and video-text datasets are usually smaller than those of text datasets.
To utilize the rich vocabulary of text dataset, our \methodname{} transfers the knowledge of pretrained multi-modal model to a student language model with a large text dataset.

\subsection{Visually-Grounded Language Learning}
A series of works explore using visual information to aid language understanding and generation tasks including co-reference resolution \cite{kong2014you, christie2016resolving},
machine translation \cite{zhang2019neural}, bilingual lexicon learning \cite{kiela2015visual}, and multi-modal contrastive learning \cite{li2020unimo}.
Vokenization \cite{tan2020vokenization} proposes the visually-supervised language model, which is closest to our work. 
Vokenization proposes to supervise a language model to predict a visualized token, called `voken' for each input text token. Vokens are obtained by a contextualized token-to-image matching model, pretrained on a MS COCO image captioning dataset \cite{chen2015microsoft}.
In this work, we experiment with alternative objectives which avoid the approximation error from finite voken assignments. In addition, we use \howtohundredM{} \cite{miech2019howto100m} video dataset, which provides a more diverse vocabulary as well as richer world commonsense and physical reasoning knowledge.

\section{Video-Language Knowledge Distillation}

\subsection{Method Overview}
\label{sec:setup}
\begin{figure*}[t]
  \centering
  \includegraphics[width=0.88\textwidth]{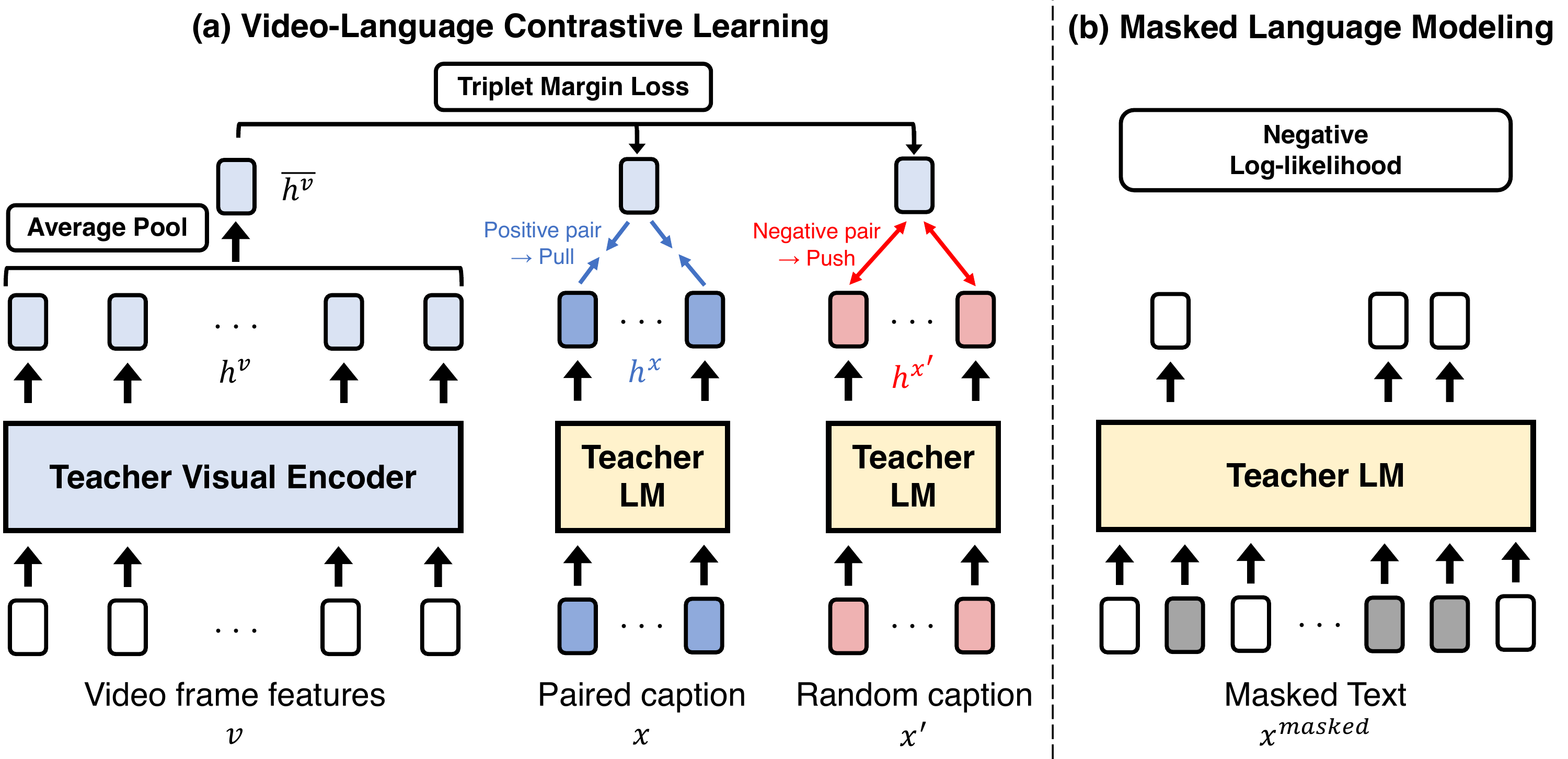}
  \caption{Cross-modal pretraining of our teacher model on a multi-modal dataset (Sec.~\ref{sec:teacher_model}). 
  We train our teacher model with (a) video-language contrastive learning 
  and (b) masked language modeling.
For video-language contrastive learning, we only illustrate the negative text samples for brevity. 
  }
  \label{fig:teacher}
\end{figure*}

\begin{figure*}[t]
  \centering
  \includegraphics[width=0.88\textwidth]{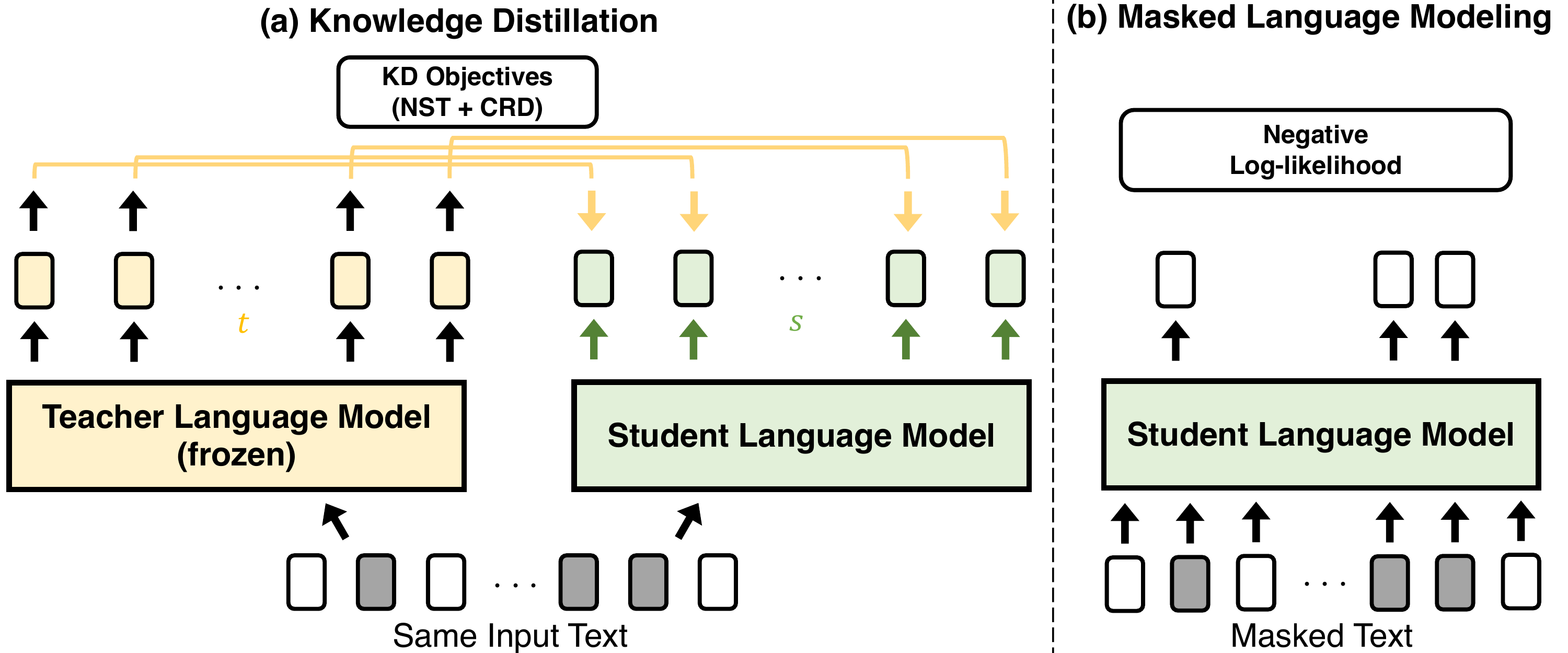}
  \caption{Illustration of our knowledge distillation from teacher language model \LMteacher{} to student language model \LMstudent{} on a text dataset (Sec.~\ref{sec:student_model}).
  We train our student model with (a) knowledge distillation objectives and (b) masked language modeling.
  }
  \label{fig:student}
\end{figure*}

$\alpha$: margin, $h^x_i$: positive text, $h^{x'}_i$: negative text,\\
$\overline{h^v}$: video representation, $\overline{h^{v'}}$: negative video representation\\
We aim to learn a better language representation with the knowledge distilled from visual information.
For this, we leverage two kinds of datasets: the aligned muti-modal 
dataset, \vldataset{}$: \{(\mathbf{x}, \mathbf{v})\}$ (e.g., \howtohundredM{} \cite{miech2019howto100m});
and the text dataset, \ldataset{}$: \{\mathbf{x}\}$ (e.g., Wikipedia),
where $\mathbf{x}$ is a sentence and $\mathbf{v}$ is a video paired with $\mathbf{x}$.
Our knowledge transfer is done in two stages:
(1) cross-modal pretraining of a teacher model, \teachermodel{}, on multi-modal data \vldataset{} (Eq.~\ref{eq:teacher})
(2) distilling the knowledge of teacher model to a student model, \studentmodel{}, on text data \ldataset{} (Eq.~\ref{eq:KD}). 
We illustrate our two-stage knowledge transfer method in Fig.~\ref{fig:teaser}.
\begin{align}
\min_{\theta^{T}} & \mathop{\mathbb{E}}_{\mathbf{x}, \mathbf{v} \sim D_{\operatorname{VL}}} \mathcal{L}^{T}(\operatorname{M}^{T}, \mathbf{x}, \mathbf{v}) \label{eq:teacher} \\
\min_{\theta^{S}} & \mathop{\mathbb{E}}_{\mathbf{x} \sim D_{\operatorname{L}}} \mathcal{L}^{\mathit{KD}}(\operatorname{M}^{T}, \operatorname{M}^{S}, \mathbf{x}) 
\label{eq:KD}
\end{align}
Our teacher model \teachermodel{} consists a language model \LMteacher{} and a visual encoder \Visteacher{}.
Both \LMteacher{} and \Visteacher{} have transformer \cite{vaswani2017attention} architectures, where \LMteacher{} takes text tokens $\mathbf{x}$ and \Visteacher{} takes video frame features $\mathbf{v}$ as inputs. 
Our student model \studentmodel{} is a transformer language model \LMstudent{} sharing the same architecture with teacher language model \LMteacher{}.
As illustrated in Fig.~\ref{fig:teaser}(a), we first train teacher models \LMteacher{} and \Visteacher{} with contrastive learning and masked language modeling.
Then, we distill the knowledge of teacher models to student model \LMstudent{} as in Fig.~\ref{fig:teaser}(b).
In the following subsections, we discuss the detailed training procedure of teacher (Sec.~\ref{sec:teacher_model}, Fig.~\ref{fig:teacher}) and student models (Sec.~\ref{sec:student_model}, Fig.~\ref{fig:student}).

\subsection{Teacher Model}
\label{sec:teacher_model}
We train our teacher model on a multi-modal dataset with two objectives, i.e.,
video-language contrastive learning
(Fig.~\ref{fig:teacher}(a))
and masked language modeling
(Fig.~\ref{fig:teacher}(b)): $\mathcal{L}^{T} = \mathcal{L}_{CT} + \mathcal{L}_\mathit{MLM}$
\footnote{In our experiments, different weights over the objectives did not significantly change the results.}

\paragraph{Architecture.} 
As shown in Figure \ref{fig:teacher}, our teacher model \teachermodel{} consists of a language encoder \LMteacher{}
and a visual encoder \Visteacher{}. 
Both \LMteacher{} and \Visteacher{} have similar
transformer architecture.\footnote{In our experiments, we use the $\text{BERT}$ architecture with two different configurations: 12 layers/768 hidden dimensions (\berttwelve{} = $\text{BERT}_\text{BASE}$) and 6 layers/512 hidden dimensions (\bertsix{}).
}
For each sentence $\mathbf{x}$, we tokenize it and append a special token \texttt{[CLS]} that represents the entire sentence following \citet{devlin2018bert}.
\LMteacher{} takes $\mathbf{x}$ and outputs contextualized representation $\mathbf{h}^x = \{\mathbf{h}^x_{\texttt{[CLS]}}, \mathbf{h}^x_{1} \cdots \mathbf{h}^x_{|x|} \}$.
For each video $\mathbf{v}$, we extract frame-level features $\mathbf{e}^{v}$ with an off-the-shelf image encoder
(see more details in Sec.~\ref{sec:videofeat}). 
Note that the parameters of the image encoder are not updated to save computation.
We feed the frame features
$\mathbf{e}^{v}=\{\mathbf{e}^{v}_1 \cdots \mathbf{e}^{v}_{|v|}\}$
to our visual encoder \Visteacher{} to get contextualized video frame features $\mathbf{h}^v = \{\mathbf{h}^v_1 \cdots \mathbf{h}^v_{|v|}\}$.
We get the final video representation $\overline{\mathbf{h}^v}$ by temporally averaging frame-level features: $\overline{\mathbf{h}^v} = \frac{1}{|v|}\sum_{i=1}^{|v|}\mathbf{h}^v_i$.
Different from \citet{tan2020vokenization}, both \LMteacher{} and \Visteacher{} parameters are trained from scratch. 

\paragraph{Video-Language Contrastive Learning.}
\label{sec:teacher_training_ct}
To learn multi-modal grounding, we use a contrastive learning objective that encourages a closer distance between representations of aligned video-text pairs than unaligned pairs, as shown in Fig.~\ref{fig:teacher}\,(a).
For each $\mathbf{x}$, we randomly sample another text $\mathbf{x}'$ from its batch with $\mathbf{x}' \neq \mathbf{x}$. 
Similarly, for each $\mathbf{v}$, we randomly sample another video $\mathbf{v}'$ from its batch with $\mathbf{v}' \neq \mathbf{v}$. 
Then, we use hinge loss $\max\{0, \alpha  - \textit{pos} + \textit{neg}\}$ on cosine similarities:
\begin{align} 
\label{eq:ct}
\mathcal{L}_{CT}(\mathbf{x}, \mathbf{x}', \mathbf{v}, \mathbf{v'}) = \sum_{i}^{|\mathbf{x}|} [ &\max \{0,
\alpha -\cos ( \mathbf{h}^x_{i}, \overline{\mathbf{h}^v} ) + \cos ( \mathbf{h}^{x'}_{i}, \overline{\mathbf{h}^v} )\}  \\
+ &\max \{0, \alpha -\cos ( \mathbf{h}^x_{i}, \overline{\mathbf{h}^v}) + \cos ( \mathbf{h}^{x}_{i}, \overline{\mathbf{h}^{v'}} )\} ] \nonumber
\end{align}
where
$\alpha$ is the margin between the similarities of a positive pair and a negative pair.
Different from previous methods~\cite{bordes2020incorporating, hessel2019unsupervised} that exploit sentence-level contrastive loss, we follow \cite{tan2020vokenization} to construct a token-level contrastive loss (triplet margin loss) that grounds the visual information to each contextualized token output.
This fine-grained contrastive loss will help the token-level knowledge distillation in Sec.~\ref{sec:kdobjective}.

\paragraph{Masked Language Modeling.}
\label{sec:teacher_training_mlm}
For better language understanding in our teacher model, we follow BERT \cite{devlin2018bert} to use masked language modeling (MLM) objective (Fig.~\ref{fig:teacher}(b)).
By replacing 15\% of tokens in $\mathbf{x}$ with a special token \texttt{[MASK]}, we obtain a masked text $\mathbf{x}^{\text{masked}}$ with the same length. The model takes $\mathbf{x}^{\text{masked}}$ as input and learns to predict the tokens by minimizing the negative log-likelihoods:
$
\mathcal{L}_\text{MLM}(\mathbf{x, \mathbf{x}^{\text{masked}}}) = -\sum_{i \in \mathrm{Mask}} \log p(\mathbf{x}_i \mid \mathbf{x}^{\text{masked}})
\label{eq:mlm}
$, where $\mathrm{Mask}$ refers to the indices of masked tokens.

\subsection{Student Model}
\label{sec:student_model}
After we train a teacher model on a multi-modal dataset,
we transfer its knowledge to a student model on a text dataset.
Following \citet{kim2016sequence}, 
we train our student model with a sum of masked language modeling and two knowledge distillation objectives, NST and CRD (see Sec.~\ref{sec:kdobjective}): 
\begin{align}
\mathcal{L^{S}}&= \mathcal{L}_{\text{MLM}} + \mathcal{L}^{\text{KD}}_{\text{NST}} + \mathcal{L}^{\text{KD}}_{\text{CRD}}
\label{eq:distill}
\end{align}

\paragraph{Architecture.}
As shown in Fig.~\ref{fig:student}\,(a), our student model \studentmodel{} is a language model \LMstudent{} with the same transformer architecture as the teacher language model \LMteacher{}. We train \LMstudent{} from scratch.
Following previous works \cite{chen2020simple, grill2020bootstrap}, we introduce a multi-layer perceptron (MLP) distillation head on top of the last hidden states of \LMstudent{}.
In our ablation study
in appendix,
we find that adding a distillation head slightly improves the distillation performance.

\subsection{Knowledge Distillation Objectives}
\label{sec:kdobjective}
We next describe the knowledge distillation (KD) objectives used in transferring knowledge from the teacher model \LMteacher{} (Sec.~\ref{sec:teacher_model}) to this student model \LMstudent{}. 
Note that the weights of the teacher model \LMteacher{} are frozen during the knowledge distillation process since the teacher model should not be affected by the student model's performance.
Following \citet{kim2016sequence},
we use the knowledge distillation objective
combined with the MLM objective (Fig.~\ref{fig:student}\,(b)).
Concretely, we use the same input text mask for MLM and KD objectives.
While we calculate the MLM loss only on masked positions, we calculate KD losses using all hidden states following \citet{clark2020electra} (Fig.~\ref{fig:student}).

We study the following KD objectives:
Soft-label \cite{hinton2015distilling},
L2 Regression \cite{ba2013do},
Neuron Selectivity Transfer (NST) \cite{huang2017like}, 
Contrastive Representation Distillation (CRD) \cite{tian2019contrastive},
and Vokenization \cite{tan2020vokenization}.
In our experiments comparing different KD objectives (Table~\ref{tab:ablation_kdloss}), 
NST and CRD perform best, while the combination of them improved the performance even further.
Therefore, we propose NST+CRD for our cross-modal knowledge distillation objective.

\vspace{3pt}
\noindent\textbf{Soft Label:}
\citet{hinton2015distilling} proposed a knowledge transfer method by taking a teacher model prediction with temperature scaling as a `soft label'.
We minimize cross-entropy between $P^T(y|x)$ and $P^S(y|x)$, i.e., the word output probabilities of \LMteacher{} and \LMstudent{} given the input text $\mathbf{x}$ respectively:
\begin{align}
\mathcal{L}^{\text{KD}}_\text{soft-label}(\mathbf{x}) = - \sum^{|\mathbf{x}|}_{i=1} \sum_{z \in Z} P^T(y_i = z | \mathbf{x}) \log P^S(y_i = z | \mathbf{x})
\end{align}
where
$Z$ is the word vocabulary.
Following \citet{hinton2015distilling}, we divide the softmax logits of \LMteacher{} and \LMstudent{} by a temperature parameter $\tau=2.0$.
Note that for soft-label KD, we reuse the LM head, instead of learning an additional distillation head.

\vspace{3pt}
\noindent\textbf{L2 Regression}: 
Following \citet{ba2013do} which uses feature regression for KD, we minimize the squared L2 distance between $\mathbf{s}(\mathbf{x})$ and $\mathbf{t}(\mathbf{x})$, the last hidden states of \LMteacher{} and \LMstudent{} given input text $\mathbf{x}$:
\begin{align}
\mathcal{L}^{\text{KD}}_{\operatorname{Regression}}(\mathbf{x}) = \sum^{|\mathbf{x}|}_{i=1} \left\|\mathbf{s}(\mathbf{x})_i - \mathbf{t}(\mathbf{x})_i\right\|_{2}^{2}
\end{align}

\vspace{3pt}
\noindent\textbf{Neuron Selectivity Transfer (NST)}:
\label{sec:NST}
NST \cite{huang2017like} is a KD method that transfers heatmap like spatial activation patterns of teacher neurons to student neurons.
We transfer the sequential activation patterns of $\mathbf{t}(\mathbf{x}) \in \mathbb{R}^{|\mathbf{x}| \times d}$ to $\mathbf{s}(\mathbf{x}) \in \mathbb{R}^{|\mathbf{x}| \times d}$,
where $\mathbf{t}(\mathbf{x})$ and $\mathbf{s}(\mathbf{x})$ are the last hidden states of \LMteacher{} and \LMstudent{} given input text $\mathbf{x}$, and $d$ is the hidden state dimension (\# neurons).
Following \citet{huang2017like}, we use the squared maximum mean discrepancy (MMD) \cite{JMLR:v13:gretton12a} with kernel trick to measure the distance between the activation patterns of student neurons $\{\mathbf{s}(\mathbf{x})_{{*},i}\}^d_{i=1}$ and teacher neurons $\{\mathbf{t}(\mathbf{x})_{{*}, j}\}^d_{j=1}$:
\begin{align}
\operatorname{MMD}^{2}(\mathbf{x})
=&\frac{1}{d^2} \sum_{i=1}^{d} \sum_{i^{\prime}=1}^{d} k\left[\mathbf{s}(\mathbf{x})_{{*},i}; \mathbf{s}(\mathbf{x})_{{*}, i^{\prime}}\right]
+\frac{1}{d^{2}} \sum_{j=1}^{d} \sum_{j^{\prime}=1}^{d} k\left[\mathbf{t}(\mathbf{x})_{{*}, j}; \mathbf{t}(\mathbf{x})_{{*}, j^{\prime}}\right] \nonumber \\
&-\frac{2}{d^2} \sum_{i=1}^{d} \sum_{j=1}^{d} k\left[\mathbf{s}(\mathbf{x})_{{*}, i}; \mathbf{t}(\mathbf{x})_{{*}, j}\right]
\end{align}
where we use Gaussian kernel $k[\mathbf{s}; \mathbf{t}]=\exp \left(-\frac{\|\mathbf{s}-\mathbf{t}\|_{2}^{2}}{2 \sigma^{2}}\right)$ with $\sigma=1$.
We transfer the teacher activation patterns to the student by minimizing squared MMD: $\mathcal{L}^{\text{KD}}_{\text{NST}}(\mathbf{x}) = \operatorname{MMD}^2(\mathbf{x})$

\vspace{3pt}
\noindent\textbf{Contrastive Representation Distillation (CRD)}: 
\label{sec:crd}
CRD \cite{tian2019contrastive} is a KD objective which maximizes the mutual information between the teacher and student representations with contrastive learning.
Let's denote $\textbf{s} \in {S}$ and $\textbf{t} \in {T}$ as student and teacher representations given $\textbf{x}$.
We are given 1 positive pair (drawn from the joint distribution) for every $N$ (batch size) negative pairs (drawn from the product of marginals; independent randomly drawn inputs from $T$ and $S$). 
Following \cite{tian2019contrastive}, we maximize the lower bound of mutual information between $\textbf{s}$ and $\textbf{t}$ by minimizing the following term:
\begin{align} 
\mathcal{L}^{\text{KD}}_{\text{CRD}}(\textbf{x}) &= -\mathbb{E}_{q(\mathbf{s}, \mathbf{t} \mid \text{postive})}[\log h(\mathbf{s}, \mathbf{t})]
-N \cdot \mathbb{E}_{q(\mathbf{s}, \mathbf{t} \mid \text{negative})}[\log (1-h(\mathbf{s}, \mathbf{t}))]  \\
h(\mathbf{s}, \mathbf{t}) &=\frac{\exp{(f_1(\mathbf{s})^{\top} f_2(\mathbf{t}))}}{\exp{(f_1(\mathbf{s})^{\top} f_2(\mathbf{t}))+\frac{N}{M}}} \nonumber
\end{align}
where $M$ is the cardinality of the dataset,
$f_1, f_2$ are learned linear layers followed by $L2$ normalization, which map the student and teacher representations
into a same feature space.
Since a large $N$ leads to a tight mutual information lower bound,
following \cite{tian2019contrastive}, we implement a memory buffer that stores the latent features of each data sample computed from previous batches. Therefore, during training we can efficiently retrieve a large number of negative samples from the memory buffer.
Note that since CRD is based on contrastive learning, it is the only KD objective where student and teacher language models can take different inputs.

\noindent\textbf{Vokenization}:
Vokenization~\cite{tan2020vokenization} could be viewed as a knowledge distillation method, where token-level text-to-image retrieval results (called `vokens') of a multi-modal matching model are used as labels for a student language model.
For the i-th input token $\mathbf{x}_i$,
we calculate cosine similarity between the i-th teacher language model hidden state $\mathbf{t}(\mathbf{x})_i$ and a video feature
$\mathbf{v}$.
Out of 30K pre-selected videos, we select a video that maximizes cosine similarity and use it as the voken for $\mathbf{x}_i$.
By denoting the voken of $\mathbf{x}_i$ as $\mathbf{voken}_{i}$, we formulate our vokenization-based KD objetive as:
\begin{align}
\mathcal{L}^{\text{KD}}_\text{Voken}(\mathbf{x}) = - \sum^{|\mathbf{x}|}_{i=1} \log P^S_{\text{voken}}(y_i = \mathbf{voken}_i | \mathbf{x})
\end{align}
where $P^S_{\text{voken}}(y|\mathbf{x})$ is the voken classification probabilities of \LMstudent{} given input text $\mathbf{x}$.
We experiment with vokenization-based KD by retrieving vokens from images and videos (see Table~\ref{tab:ablation_vokenkd} of Sec.~\ref{sec:ablation}).
Note that vokenization suffers from approximation error; it's hard to cover diverse textual concepts with 30K vokens. This motivates us to experiment with different `soft' KD objectives described in this section (see Table~\ref{tab:ablation_kdloss} of Sec.~\ref{sec:ablation}).
\section{Experimental Setup}
\label{sec:expsetup}

\subsection{Datasets}
\label{sec:datasets}

\noindent\textbf{Video-Text Dataset.} \; 
We use \howtohundredM{} \cite{miech2019howto100m} for cross-modal pretraining of our teacher model (Sec.~\ref{sec:teacher_model}).
\howtohundredM{} has 1.22M videos totaling 136M video clips with total duration of 134,472 hours describing over 23K different visual tasks. There are 138M captions, 568M tokens with 633K distinct tokens.

\noindent\textbf{Text Pretraining Dataset.} \;
To transfer the knowledge from our teacher language models to student language models (Sec.~\ref{sec:student_model}),
we follow \citet{tan2020vokenization} to use English Wikipedia.
For ablation studies (Sec.~\ref{sec:ablation}), we use \wikihundredthree{}~\cite{merity2016pointer}, a widely used subset of English Wikipedia.
There are 2.9B tokens and 120M sentences in English Wikipedia, and 111M tokens and 4.2M sentences in \wikihundredthree{}.

\noindent\textbf{Text Downstream Dataset.} \;
Following \citet{tan2020vokenization},
we finetune our models on GLUE \cite{wang2018glue}, SQuAD \cite{rajpurkar2016squad} 1.0 and SQuAD2.0 \cite{rajpurkar2018know}, and SWAG \cite{zellers2018swag} to assess the pretraining performance.
Since some smaller tasks in GLUE are reported as unstable in recent papers \cite{dodge2020fine},
we evaluate on the four largest datasets of GLUE: SST-2 \cite{cer2017semeval}, QNLI \cite{rajpurkar2016squad}, QQP \cite{iyer2017qqp}, and MNLI \cite{williams2017broad}. 
In addition, we also evaluate our models on the GLUE diagnostics  \cite{wang2018glue}, PIQA \cite{bisk2020piqa}, and TRACIE \cite{zhou-etal-2021-temporal} to measure its linguistic knowledge, physical reasoning, and temporal reasoning abilities. 
 
\subsection{Video Feature Representations}
\label{sec:videofeat}

Following \citet{miech2019howto100m},
we encode video features by concatenating features from a 2D frame-level image encoder and a 3D video encoder in channel dimension.
Note that the parameters for 2D image encoder and 3D video encoder are not updated.

For the 2D image encoder, we sample video frames by 1fps (frame/second). The 2D image encoder outputs features for each frame individually.
We experiment with ResNet-152 \cite{he2016deep} pretrained on ImageNet-1K~\cite{deng2009imagenet} and CLIP \cite{radford2021learning} image encoder (ViT-B/32 \cite{dosovitskiy2020image}). In contrast to conventional image encoders trained with image label classification, the CLIP image encoder is trained to match a corresponding natural language description by large-scale contrastive learning.
We discuss if this natural language supervision can help our cross-modal KD in
Sec.~\ref{sec:downstream}.

For the 3D video encoder, we use 3D-ResNeXt-152\footnote{\url{https://github.com/kenshohara/3D-ResNets-PyTorch}}  \cite{Xie_2018_ECCV, hara3dcnns, kataoka2020would} trained from a combination of publicly available datasets: ActivityNet \cite{caba2015activitynet}, Kinetics \cite{kay2017kinetics}, UCF-101 \cite{soomro2012ucf101}, and HMDB-51 \cite{kuehne2011hmdb}.
The 3D video encoder processes 24fps videos with 3D convolution and yields features at 1.5fps.
Then we sub-sample the features to 1fps to match the frame rate of 2D image encoder.

\subsection{Implementation Details}

For the student distillation head, we use a two-layer MLP with ReLU activation.
For both student and teacher language models,
following previous works~\cite{liu2019roberta, conneau2019unsupervised, tan2020vokenization}, we truncate input text that is longer than 128 tokens.
We truncate videos features that are longer than 512 frames.
We use an AdamW~\cite{kingma2014adam} optimizer with learning rate 2e-4 and weight decay \cite{loshchilov2017decoupled} of 0.01. 
We reserve 10K samples of the \howtohundredM{} dataset as validation data.
We train the teacher model until it converges on validation data. 
For downstream tasks, we report the results on the validation sets. We train 3 epochs with a learning rate of 1e-4 and a batch-size of 32 for all downstream tasks. 
We use hinge loss margin $\alpha=1.0$ for $\mathcal{L}_{\operatorname{CT}}$ (Eq.~\ref{eq:ct}).
We implement our models with PyTorch 1.5 \cite{paszke2017automatic} and train them with Nvidia GeForce RTX 2080ti GPUs.
For teacher pretraining, we use 4 GPUs for \berttwelve{} and \bertsix{} models for 7 days and 2.5 days respectively.
For knowledge distillation, we use 4 GPUs for \berttwelve{} and \bertsix{} models for 10 days and 3 days respectively.

\section{Results and Analysis}

\begin{table*}[t]
\vspace{-4pt}
\caption{
Cross-modal knowledge distillation results of \berttwelve{} student language model on 7 downstream NLU tasks.
In the first block,
we include the image-based vokenization (Img-Voken) and its text-only pretrained baseline performance from \citet{tan2020vokenization}.
In the second block,
we compare our cross-modal KD method (NST+CRD) to video-based vokenization (Vid-Voken) and a text-only pretrained baseline.
$^\dagger$EM refers to `Exact Match'.
}
\centering
\vspace{-5pt}
\scalebox{0.8}{
\begin{tabular}{lcccccccc}
\toprule
 & SST-2 & QNLI & QQP & MNLI & SQuAD v1.1 & SQuAD v2.0 & SWAG & Avg.\\
 & Acc & Acc & Acc & Acc & EM$^\dagger$ & EM & Acc & \\
\midrule

\berttwelve{} \cite{tan2020vokenization} & 89.3 & 87.9 & 83.2 & 79.4 & 77.0 & 67.7 & 65.7 & 78.6 \\
+ KD (Img-Voken) \cite{tan2020vokenization}  & 92.2 & 88.6 & 88.6 & 82.6 & 78.8 & 68.1 & 70.6 & 81.4 \\

\midrule

\berttwelve{} & 89.0 & 88.0 & 86.2 & 79.2 & 77.2 & 68.0 & 65.0 & 78.9 \\
+ KD (Vid-Voken) w/ ResNet & 93.4 & 89.2 & 88.7 & 83.0 & 78.9 & 68.7 & 70.0 & 81.7 \\
+ KD (Vid-Voken) w/ CLIP & 94.1 & \textbf{89.8} & 89.0 & 83.9 & 79.2 & 68.6 & 71.6 & 82.3 \\
+ KD (NST+CRD) w/ ResNet & 94.2 & 89.3 & 89.7 & 84.0 & 79.0 & \textbf{68.9} & 71.8 & 82.4 \\
+ KD (NST+CRD)  w/ CLIP & \textbf{94.5} & 89.6 & \textbf{89.8} & \textbf{84.2} & \textbf{79.6} & 68.7 & \textbf{72.0} & \textbf{82.6} \\

\bottomrule
\end{tabular}
}
\label{tab:downstream}
\vspace{-4pt}
\end{table*}

\subsection{Primary Downstream Task Results}
\label{sec:downstream}

In the first block of Table~\ref{tab:downstream}, we include the image-based vokenization (Img-Voken) and their text-only pretrained baseline from \citet{tan2020vokenization}.\footnote{
Vokenization uses pretrained BERT checkpoint for its `teacher' (vokenizer) model but we train our teacher language model fully from scratch.
}
Given our reproduced text-only baseline shows a similar average performance (78.9 vs 78.6), our student models distilled from NST+CRD are much better (82.6 vs 81.4).
We discuss the comparison between video-based and image-based KD in detail in the following ablation study in comparison to vokenization
(Table~\ref{tab:ablation_vokenkd}).

In the second block of Table~\ref{tab:downstream}, we compare our proposed cross-modal KD method (NST+CRD) to video-based vokenization (Vid-Voken) and a non-KD baseline (\berttwelve{}) which is only pretrained on text.
We can see both cross-modal KD methods (i.e., KD and Vid-Voken) significantly outperform the text-only baseline across all 7 downstream tasks.
We also experiment with different 2D frame encoders (Sec.~\ref{sec:videofeat}): ResNet and CLIP.
For both Vid-Voken and NST+CRD, we observe CLIP further improves the performance results over ResNet,
indicating using a strong visual encoder helps the teacher training and thus benefits the knowledge distillation.

\subsection{Ablation Studies}
\label{sec:ablation}

In this section, we conduct a comprehensive ablation study to show the effectiveness of our proposed methods.
For all ablation experiments, we use \bertsix{} architecture for student and teacher language models.
We use ResNet-152 for 2D frame encoder and 3D-ResNeXt-152 for 3D frame encoder (Sec.~\ref{sec:videofeat}).
\wikihundredthree{} \cite{merity2016pointer} is used for student model training.
We also perform ablation experiments on the effect of additional distillation head in appendix.

\begin{wraptable}{r}{80mm}
\centering
\caption{
Text-only pretraining results of \bertsix{} pretrained on \wikihundredthree{}, \howtohundredM{} captions, and no-pretrain baseline.
}
\label{tab:ablation_datasets}
\scalebox{0.80}{
\begin{tabular}{lcccc}
\toprule
Pretrained on & SST-2 & QNLI & QQP & MNLI \\
\midrule
No-Pretrain & 79.6 & 61.5 & 72.7 & 61.6 \\
\wikihundredthree{} (Formal language) & \textbf{88.8} & \textbf{84.9} & \textbf{85.3} & \textbf{77.4} \\
\howtohundredM{} (ASR captions) & 83.3 & 78.5 & 83.7 & 71.5\\
\bottomrule
\end{tabular}
}
\vspace{3mm}
\end{wraptable}

\paragraph{Text-only Pretraining.}
Our cross-modal KD improves the performance on downstream NLU tasks significantly (Sec.~\ref{sec:downstream}).
Where does the improvement come from, video or text?
To answer this question, 
we conduct text-only pretraining of \bertsix{} on \wikihundredthree{} text (111M tokens), \howtohundredM{} captions (568M tokens) and compare them to a no-pretrain baseline.
In Table~\ref{tab:ablation_datasets}, 
while both pretrained models improve the performance over the no-pretrain baseline, \wikihundredthree{}-trained model outperforms \howtohundredM{}-trained model (which has more tokens) significantly.
This indicates that our KD methods improve NLU performance because of multimodal grounding, instead of just the larger corpus.

\begin{wraptable}{r}{75mm}
\centering
\caption{Ablation results showing the effect of the teacher model's training objectives. NST is used for knowledge distillation.}
\label{tab:teacher_objective}
\scalebox{0.80}{
\begin{tabular}{lcccc}
\toprule
 & SST-2 & QNLI & QQP & MNLI \\
\midrule
\bertsix{} & 88.8 & 84.9 & 85.3 & 77.4 \\
+KD from ${T}^{\mathit{MLM}}$ & 88.1 & 83.1 & 85.6 & 77.4 \\
+KD from ${T}^{\mathit{CT}}$ & 88.9 & \textbf{85.2} & 86.2 & 77.5 \\
+KD from ${T}^{\mathit{MLM}+\mathit{CT}}$ & \textbf{91.1} & 85.0 & \textbf{87.4} & \textbf{78.4} \\
\bottomrule
\end{tabular}
}
\vspace{3mm}
\end{wraptable}

\paragraph{Effect of Teacher Training Objectives.}
We here analyze the teacher training objectives by comparing the corresponding distilled student model results.
In Table~\ref{tab:teacher_objective}, the teacher model trained solely with MLM (+KD from ${T}^\mathit{MLM}$) does not significantly change the student model performance.
At the same time, the teacher model trained with only visual supervision, i.e., contrastive objective (+KD from ${T}^{CT}$), improves the result.
This illustrates the motivation to perform knowledge transfer from a visually supervised MLM model.
Lastly, combining the MLM and the contrastive objective (+KD from ${T}^\mathit{MLM+CT}$) in teacher model training shows the best student results.

\begin{wraptable}{r}{75mm}
\centering
\caption{Comparison of pretraining on text, video, both (Two-stage PT), and our VidLanKD.
}
\label{tab:student_teacher}
\scalebox{0.85}{
\begin{tabular}{lcccc}
\toprule
Model & SST-2 & QNLI & QQP & MNLI \\
\midrule
Text PT & 88.8 & 84.9 & 85.3 & 77.4 \\
Video PT& 84.0 & 78.9 & 84.2 & 73.1\\
Two-Stage PT & 90.3 & \textbf{85.0} & 87.2 & 76.9\\
\methodname{} &  \textbf{91.1} & \textbf{85.0} & \textbf{87.4} & \textbf{78.4} \\
\bottomrule
\end{tabular}
}
\vspace{2mm}
\end{wraptable}

\paragraph{Two-stage PT vs. Cross-modal KD.}
In Table~\ref{tab:student_teacher},
we compare two-stage pretraining with a single model to our proposed cross-modal KD approach.
For single model baselines, we use text-only (MLM on \wikihundredthree{}), video-only (MLM+CT on \howtohundredM{}), and two-stage (video-then-text) pretraining.
While the two-stage pretraining shows better results than the text/video-only pretraining,
our \methodname{} outperforms all baselines on GLUE tasks, especially on SST-2 and MNLI.

\begin{wraptable}{r}{75mm}
\caption{Ablation of knowledge distillation objectives.
}
\label{tab:ablation_kdloss}
\centering
\scalebox{0.75}{
    \begin{tabular}{lcccc}
    \toprule
     & SST-2 & QNLI & QQP & MNLI \\
    \midrule
    \bertsix{} & 88.8 & 84.9 & 85.3 & 77.4 \\
    +KD-Soft label & 87.2 & 84.4 & 86.4 & 76.6\\
    +KD-Regression & 88.8 & 84.8 & 87.1 & 78.1\\
    +KD-Vid Voken & 89.7 & 85.5 & 86.5 & 77.8 \\
    +KD-NST & \textbf{91.1} & 85.0 & \textbf{87.4} & \textbf{78.4} \\
    +KD-CRD & 90.0 & \textbf{85.5} & 87.3 & 78.3 \\
    \midrule
    +KD-NST+CRD & \textbf{91.5} & \textbf{85.8} & \textbf{87.4} & \textbf{78.7}\\
    \bottomrule
    \end{tabular}
}
\end{wraptable}

\paragraph{KD Objectives Comparison.}
\label{sec:KDcomparison}

In Table \ref{tab:ablation_kdloss}, we compare different knowledge distillation objectives introduced in Sec.~\ref{sec:kdobjective}.
The student models trained with NST \cite{huang2017like} and CRD \cite{tian2019contrastive} show the best finetuning performance on downstream tasks. 
When combining NST and CRD, performance further improves with marginal additional computation cost, hence we propose to use NST+CRD for our cross-modal knowledge distillation.

\begin{wraptable}{r}{75mm}
\centering
\caption{Comparison between vokenization (Voken) and NST with image and video-level supervision.}
\label{tab:ablation_vokenkd}
\scalebox{0.75}{
\begin{tabular}{lcccc}
\toprule
 & SST-2 & QNLI & QQP & MNLI \\
\midrule
\bertsix{} & 88.8 & 84.9 & 85.3 & 77.4 \\
+KD-Voken (Image) & 89.3 & 84.4 & 86.0 & 77.5 \\
+KD-NST (Image) & 88.9 & 85.0 & 86.3 & 77.2\\
\midrule
+KD-Voken (Video) & 89.7 & \textbf{85.5} & 86.5 & 77.8 \\
+KD-NST (Video) & \textbf{91.1} & 85.0 & \textbf{87.4} & \textbf{78.4} \\
\bottomrule
\end{tabular}
}
\end{wraptable}

\paragraph{Comparison to Vokenization.}
In Table~\ref{tab:ablation_vokenkd}, we compare NST \cite{huang2017like} and Vokenization \cite{tan2020vokenization} in both image and video-level teacher model supervision.
For video-level supervision, we provide our visual encoder with the whole video features (Sec.~\ref{sec:videofeat}).
For image-level supervision, we provide our visual encoder only with 2D features of the middle frame for each video clip.
With image-level supervision (first block), Vokenization and NST show comparable performance.
However, with video-level supervision (second block), NST outperforms Vokenization on 3 out of 4 tasks.
The gap in the video domain might come from voken approximation error, where each image or video input is approximated with one of 30K predefined vokens.
Since videos usually contain more diverse contents than images, the voken approximation error would be amplified in video-level supervision, whereas our NST distillation avoids this issue.

\subsection{Analyzing the Knowledge Learned from Video}
\label{subsec:knowledge_analysis}

In this subsection, we analyze the knowledge that our language models learn from video via cross-modal knowledge distillation.
To measure linguistic knowledge and physical/temporal reasoning ability,
we show results of our models on the GLUE diagnostics \cite{wang2018glue}, the Physical Interaction Question Answering (PIQA) \cite{bisk2020piqa}, and TRACIE \cite{zhou-etal-2021-temporal}.
In addition, we visualize the learned multi-modal grounding ability of our model with text-to-video retrieval.

\vspace{3mm}
\begin{table*}[h]
\centering
  \caption{
  Finetuning performance on GLUE diagnostics~\cite{wang2018glue},
  PIQA~\cite{bisk2020piqa} and TRACIE~\cite{zhou-etal-2021-temporal} datasets, which measure the linguistic knowledge, physical and temporal reasoning capabilities of language models, respectively.
  }
  \label{tab:knowledge_analysis}
  \scalebox{0.95}{
  \begin{tabular}{lcccccc}
    \toprule
    & \multicolumn{4}{c}{GLUE diagnostics} & \multirow{2}{*}{PIQA} & \multirow{2}{*}{TRACIE} \\
    \cmidrule{2-5}
     & Lexicon & Predicate & Logic & Knowledge\\
    \midrule
    \bertsix{} & 53.0 & 64.2 & 44.5 & 44.0 & 56.9 & 63.4 \\
    + KD-NST  & 53.3 (+0.3) & 63.7 (-0.5) & 44.8 (+0.3) & 48.6 \textbf{(+4.6)}  & 60.0 (\textbf{+3.1}) & 66.7 (\textbf{+3.3}) \\
    \bottomrule
  \end{tabular}
  }
\end{table*}

\paragraph{Linguistic Knowledge.}
GLUE diagnostics dataset \cite{wang2018glue} evaluates sentence understanding through natural language inference (NLI) problems.
The dataset consists of sentence pairs labeled with their entailment relations (entailment, contradiction, or neutral) in both directions and tagged with a set of entailment labels. 
Each example in the dataset is labeled with 4 categories of linguistic phenomena:
(1) lexical semantics,
(2) predicate-argument structure,
(3) logic,
and (4) knowledge (including common sense). 
In Table~\ref{tab:knowledge_analysis}, we compare the baseline language model 
(\bertsix{} pretrained on \wikihundredthree{}) to our NST-distilled model.
We finetune the models on MNLI \cite{williams2017broad} that has the same format and test on GLUE diagnostics.
We observe a large gain on the knowledge category (which involves common sense and external world knowledge) while there are no significant differences on other categories. This suggests that our student model learns the external, grounded world knowledge in the teacher model and the video-text dataset.

\paragraph{Physical and Temporal Reasoning.}
PIQA~\cite{bisk2020piqa} is a question answering dataset evaluating physical interactions and commonsense reasoning.
TRACIE~\cite{zhou-etal-2021-temporal} is a temporal reasoning benchmark on implicit events, which are not mentioned explicitly in natural language text but can be inferred from it.
In Table~\ref{tab:knowledge_analysis}, our \bertsix{} distilled with NST significantly outperform the text-only pretrained baseline on both benchmarks.
The finding suggests (consistent with the GLUE diagnostics findings above) that video knowledge distillation also helps improve the physical and temporal reasoning capabilities of the language model. 
See appendix for the more detailed discussion on the PIQA and TRACIE experiment.

\begin{figure*}[t]
  \centering
  \includegraphics[width=0.95\textwidth]{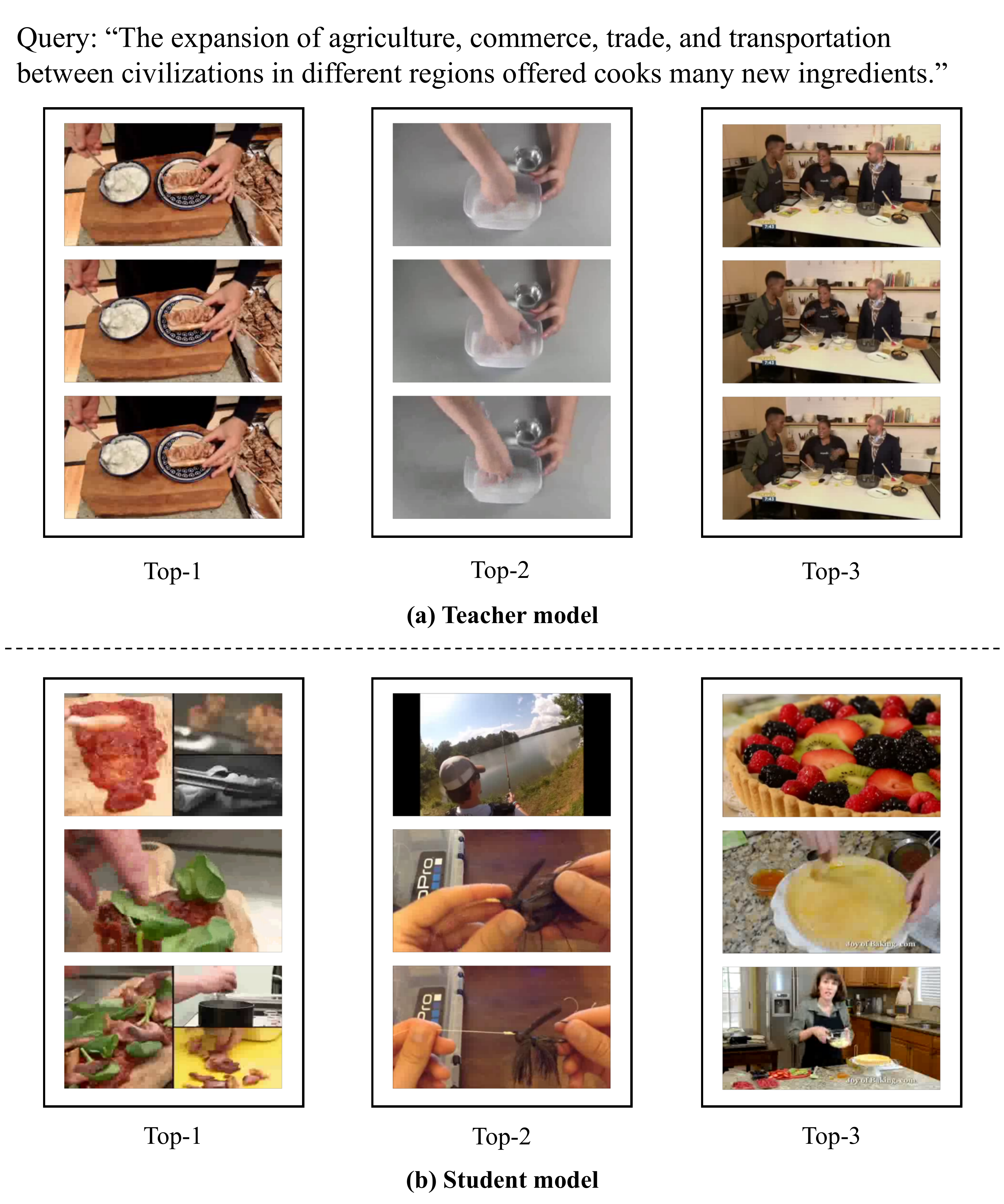}
  \caption{Text-to-video retrieval results from our teacher and student language model.}
  \label{fig:visualization_1}
\end{figure*}
\vspace{4mm}

\paragraph{Visualization: Text-to-Video Retrieval.}
\label{sec:visualization}
Our teacher language model learns to predict a corresponding video feature for each input text token (Sec.~\ref{sec:teacher_model}), and our student language model tries to follow the teacher's prediction.
To visualize the learned multi-modal grounding, we experiment with text-to-video retrieval using our teacher and student language models.
In Fig.~\ref{fig:visualization_1},
we provide the top 3 text-to-video retrieval results from teacher and student models using same input sentences.
We observe that, in many cases, both our teacher and student model can retrieve video clips that are semantically aligned to input text.
Note that this is a surprising and positive result because our student model does not see any visual input during its training (Sec.~\ref{sec:student_model}), which means the multi-modal grounding ability is learned solely from the knowledge distillation on text dataset.
See appendix for more text-to-video retrieval results and implementation details.

\section{Conclusion}

We introduce \methodname{}, a novel cross-modal knowledge distillation method to help general language understanding.
Our teacher model is first trained on a video-text dataset, and then we transfer its knowledge to a student language model with a text dataset.
Via the distillation objectives and video-text datasets, our method overcomes the limitations of the recent vokenization method.
We empirically demonstrate that our \methodname{} improves on several NLU tasks over models trained by pure-text or vokenization.
We conduct comprehensive ablation analysis to show the effectiveness of each proposed component.
We also illustrate the linguistic knowledge and physical/temporal commonsense reasoning learned from videos, and visualize our model's multi-modal grounding ability.

\section*{Acknowledgments}
We thank the reviewers for their helpful comments.
We thank Yixin Nie and Gabriel Ilharco for useful dataset suggestions.
This work was supported by ARO-YIP Award W911NF-18-1-0336, DARPA MCS Grant N66001-19-2-4031, DARPA KAIROS Grant FA8750-19-2-1004, Google Focused Research Award, and Bloomberg Data Science Ph.D. Fellowship. The views, opinions, and/or findings contained in this article are those of the authors and not of the funding agency.


{\small
\bibliographystyle{acl_natbib}
\bibliography{bib}
}

\appendix

In this appendix,
we start with describing the experimental setup details (Sec.~\ref{sup_sec:expsetup}).
We provide
ablation study on distillation head (Sec.~\ref{sup_sec:ablation}),
details of physical (Sec.~\ref{sup_sec:physical_reasoning}) and temporal (Sec.~\ref{sup_sec:temporal_reasoning}) reasoning analysis,
details of text-to-video visualization (Sec.~\ref{sup_sec:visualization}),
and broader impacts and limitations (Sec.~\ref{sup_sec:limitations}).

\section{Experimental Setup}
\label{sup_sec:expsetup}

\paragraph{Video Voken Sampling.}
To ensure the diversity of video vokens, we first select a video for each of 23K visual task.
For the remaining 7K vokens, we randomly select 7K visual tasks, then select a video from each visual task.
Each sampled video from 30K has on average around 100 clips. We select one clip from each video with length ranging from 1 to 20 seconds.

\section{Additional Distillation Head}
\label{sup_sec:ablation}

To investigate whether the additional MLP distillation head (Sec.~\ref{sec:student_model} in the main paper) affects the distillation performance, we do an ablation by conducting knowledge distillation directly on the last hidden states of student language models.
As we see in Table~\ref{tab:ablation_head}, for both NST and CRD, the performance drops on all downstream tasks when distillation heads are removed.
This finding is consistent with recent works~\cite{chen2020simple, tian2019contrastive}.

\vspace{3mm}
\begin{table*}[h]
\centering
\caption{Ablation results of additional distillation heads for student language models.
}
\label{tab:ablation_head}
\begin{tabular}{lcccc}
\toprule
 & SST-2 & QNLI & QQP & MNLI \\
\midrule
\bertsix{} & 88.8 & 84.9 & 85.3 & 77.4 \\
+KD-NST & \textbf{91.1} & 85.0 & \textbf{87.4} & \textbf{78.4} \\
+KD-CRD & 90.0 & \textbf{85.5} & 87.3 & 78.3\\
\midrule
+KD-NST (w/o head) & 89.4 (-0.7) & 84.8 (-0.2) & 86.7 (-0.7) & 77.0 (-1.4) \\
+KD-CRD (w/o head) & 88.9 (-0.1) & 85.1 (-0.4) & 86.6 (-0.7) & 77.8 (-0.5) \\
\bottomrule
\end{tabular}
\vspace{3mm}
\end{table*}

\section{Physical Reasoning Details}
\label{sup_sec:physical_reasoning}

PIQA \cite{bisk2020piqa} is a physical commonsense reasoning dataset with a format of choosing an answer among two hypotheses given context.
In Table \ref{tab:piqa}, we compare the accuracy of text-only pretraining, image-based KD and video-based KD on PIQA. While the image-based KD helps to improve accuracy from text-only pretrained model, our \methodname{} further improves the results. In Tables~\ref{tab:piqa_examples1} and  \ref{tab:piqa_examples2}, we provide PIQA question examples and related video clips from \howtohundredM{} that could help models to answer the questions.

\vspace{3mm}
\begin{table*}[h]
\centering
\caption{Performance on PIQA with teacher trained with images or video supervision. NST is used as KD objective.}
\label{tab:piqa}
\begin{tabular}{lc}
\toprule
 & TRACIE Accuracy \\
\midrule
\bertsix{} & 56.9 \\
+ Image KD & 58.9 \\
+ \methodname{} & \textbf{60.0} \\
\bottomrule
\end{tabular}
\vspace{3mm}
\end{table*}

\paragraph{Visual Grounding Improves Physical Reasoning.}
In Table~\ref{tab:piqa_examples1},
the first video clip\footnote{\url{https://www.youtube.com/watch?v=ASjB-GtyIZE}} illustrates how to fix a car cup holder that involves removing a screw with a screwdriver, which helps models to learn the action of how to remove a screw from another object.
From the second video clip\footnote{\url{https://www.youtube.com/watch?v=NQCuOKFwQ4Q}}, the model can learn from the visual of planting in soil, which helps models to identify the correct action on planting.

\begin{table*}
\small
\centering
\caption{
PIQA test set examples comparing text-only vs. video grounding.
GT stands for ground-truth labels.
Text-only refers to the text-only baseline (\bertsix{}).
Ours refers to \methodname{} student model distilled with NST objective from video supervised teacher model.
}
\label{tab:piqa_examples1}
\begin{tabular}{p{0.18\textwidth}p{0.25\textwidth}p{0.25\textwidth}P{0.04\textwidth}P{0.06\textwidth}P{0.04\textwidth}}
\toprule
Context & Hypothesis 1 & Hypothesis 2 & GT & Text-only & Ours\\
\midrule
1. to remove a screw from a board, 
& (a) place the tip of the screwdriver into the top of the screw and twist in a clockwise direction. 
& (b) place the tip of the screwdriver into the top of the screw and twist in a counter clockwise direction.
& (b) & (a) & (b)  \\
2. how to grow a plant. 
& (a) bury seed in sand and add 1 cup of water daily.	
& (b) bury seed in soil and add 1 cup of water daily.
& (b) & (a) & (b)\\
\bottomrule
\end{tabular}
\end{table*}

\paragraph{Video vs. Image Grounding.}
Videos can convey more temporal information such as actions/motions. Video captions (e.g., HowTo100M) also have a larger vocabulary coverage than image captions (e.g., CC or SBU) thus more words could be effectively grounded. Therefore, videos can provide richer visual information than images.
In Table~\ref{tab:piqa_examples2},
The first video clip\footnote{\url{https://www.youtube.com/watch?v=38FqlXKZ6LA}} illustrates how to cut wood with a band saw, which helps models to answer the question.
The second video clip\footnote{\url{https://www.youtube.com/watch?v=MMtiszBnpuc}} illustrates a brisket recipe where beef is marinated and stored in a ‘cold’ fridge, which helps our model to answer the question.

\begin{table*}
\small
\centering
\caption{
PIQA test set examples comparing video vs. image Grounding.
GT stands for ground-truth labels.
Baseline refers to our student model distilled with NST objective from image-supervised teacher model.
Ours refers to \methodname{} student model distilled with NST objective from video-supervised teacher model.
}
\label{tab:piqa_examples2}
\begin{tabular}{p{0.18\textwidth}p{0.25\textwidth}p{0.25\textwidth}P{0.04\textwidth}P{0.06\textwidth}P{0.04\textwidth}}
\toprule
Context & Hypothesis 1 & Hypothesis 2 & GT & Image KD & Ours\\
\midrule
1. how to cut wood on a band saw.
& (a) get the piece of wood you want to cut and put on your safety equipment. start the saw and cut.
& (b) start the band saw and put your wood on the top. push it through the blade and let it drop to the floor.
& (a) & (b) & (a) \\
2. how do you properly prepare a steak.
& (a) take the steak out of warm storage and let come to room temperature, generously add salt and pepper to both sides and let sit for 10 minutes.
& (b) take the steak out of cold storage and let come to room temperature, generously add salt and pepper to both sides and let sit for 10 minutes.
& (b) & (a) & (b) \\
\bottomrule
\end{tabular}
\end{table*}

\section{Temporal Reasoning Details}
\label{sup_sec:temporal_reasoning}
As described in Sec.~\ref{subsec:knowledge_analysis} in the main paper,
to measure the temporal understanding ability learned from our video-text pretraining, we fine-tune our model on TRACIE \cite{zhou-etal-2021-temporal}, a temporal reasoning benchmark on implicit events --- events that are not mentioned explicitly in natural language text but can be inferred from it.
We provide three examples from TRACIE test set in Table~\ref{tab:tracie_examples}.
As illustrated in the table, TRACIE is a textual entailment task where a model infers whether a hypothesis containing a temporal comparator $\in \{\texttt{starts},\texttt{ends}\}$ and a relation $\in \{\texttt{before},\texttt{after}\}$ corresponds to a premise.
Following \cite{zhou-etal-2021-temporal}, we use the \textit{uniform-prior} training setting which 
removes the statistical correlation between comparators and relations.
Table~\ref{tab:tracie} shows the student language model distilled with our \methodname{} (+KD-NST) outperforms the accuracy of the text-only baseline (\bertsix{}) by 3.3\%.
In the right three columns of Table~\ref{tab:tracie_examples}, we show the ground truth labels and model predictions for three examples. While our student model correctly predicts all three examples, the text-only baseline fails in the last two examples.
We conjecture that it is hard to understand the meaning of words that require temporal understanding, such as `before' and `after’, only from text. \howtohundredM{} videos consist of multiple events with corresponding ASR captions, which could help models to learn the temporal relations.

\begin{table*}[t]
\small
\centering
\caption{TRACIE test set examples. Ent. and Con. stand for Entailment and Contradiction, respectively.
GT stands for ground-truth labels.
Baseline refers to the text-only baseline (\bertsix{}).
Ours refers to our student model distilled with NST objective (+KD-NST).
}
\label{tab:tracie_examples}
\begin{tabular}{p{0.5\textwidth}p{0.17\textwidth}P{0.06\textwidth}P{0.06\textwidth}P{0.06\textwidth}}
\toprule
Context (Premise) & Hypothesis & GT & Baseline & Ours \\
\midrule
"One day, Ernie went on a walk in the park." Ernie walked by the tennis courts and saw two beautiful women playing. "He had never played tennis before, but he decided to learn." "The next day he went to the park, and the ladies were there again." "They invited him to join them, and eventually one became his wife." & Ernie bought himself a tennis racquet \textbf{ends after} the next day he went back to the park. & Con. & Con. & Con. \\
\midrule
Tim was visiting his grandparents. They didn't have wifi or fast internet. Their connection was still using dial up. Tim tried to use the internet but it was just too slow. He decided to just use his smart phone instead. & Dial up internet is not as good \textbf{starts before} Tim visit his grandparents & Ent. & Con. & Ent. \\
\midrule
Paul hates his job. Everyday at work he gets angry and says mean things to people. Paul's boss gave him a verbal warning about his attitude at work. Currently Paul is on a performance plan at work. Next month Paul will be fired. & Paul is not friendly. \textbf{starts after} Paul hat his job & Ent. & Con. & Ent. \\
\bottomrule
\end{tabular}
\end{table*}

\vspace{2mm}
\begin{table*}[h]
\centering
\caption{Performance on TRACIE \textit{uniform-prior} training setting.}
\label{tab:tracie}
\begin{tabular}{lc}
\toprule
 & TRACIE Accuracy \\
\midrule
\bertsix{} & 63.4 \\
+KD-NST & \textbf{66.7} \\
\bottomrule
\end{tabular}
\end{table*}
\vspace{3mm}

\begin{figure*}[t]
  \centering
  \includegraphics[width=0.95\textwidth]{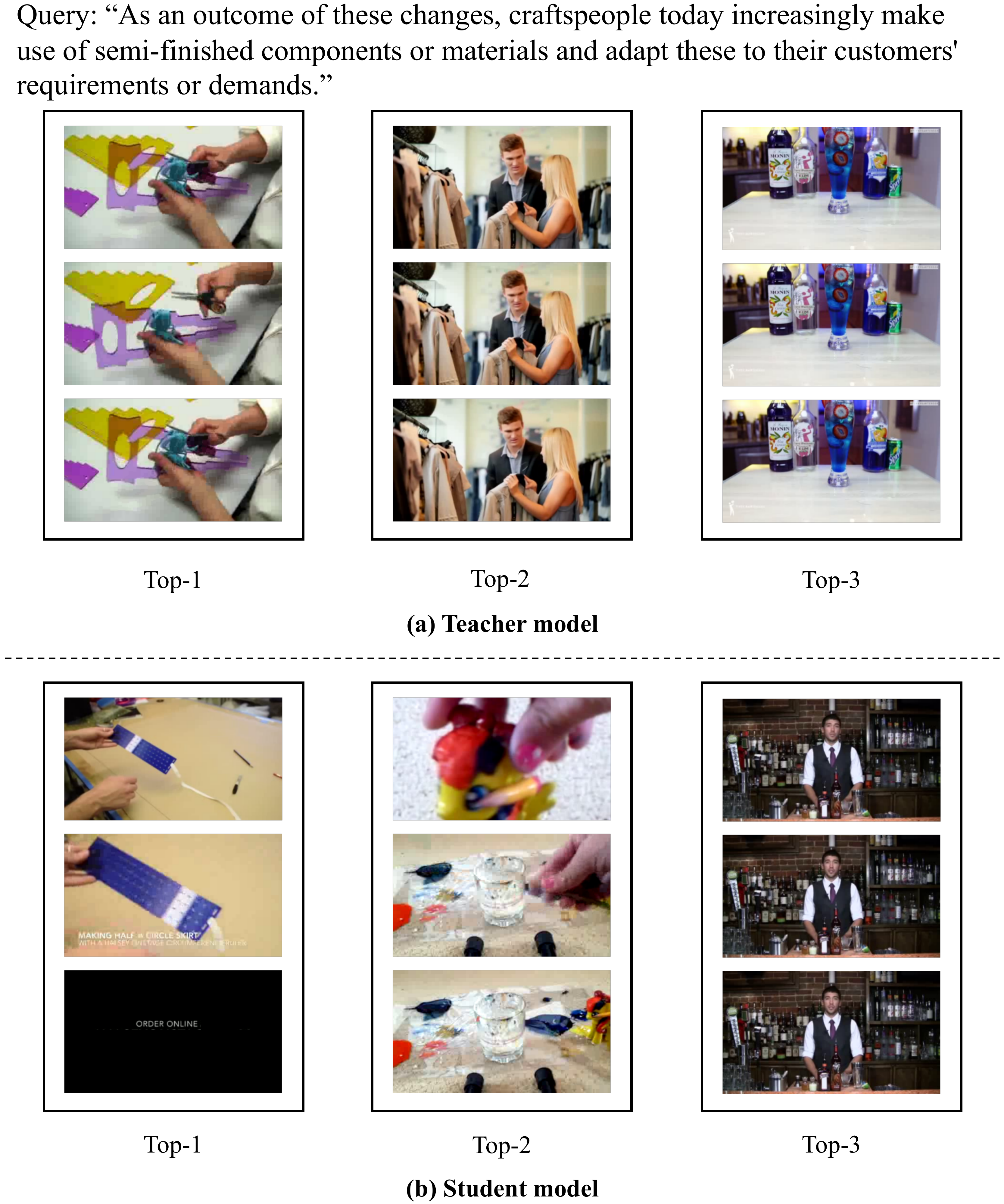}
  \caption{More text-to-video retrieval results from our teacher and student language model.}
  \label{fig:visualization_2}
\end{figure*}

\section{Visualization Details}
\label{sup_sec:visualization}

For text-to-video visualization experiment (Sec.~\ref{sec:visualization} in the main paper),
we use \bertsix{} architecture for both teacher and student (KD-NST+CRD) language models.
We sample sentences from Wikipedia and conduct text-to-video retrieval on the 60K video clips sampled from \howtohundredM{}.
For sentence feature, we use the average of the last hidden states of language models. Then we calculate the cosine similarity between the video and sentence features for relevance score.
We include more visualization results in Fig.~\ref{fig:visualization_2}.

\section{Broader Impacts and Limitations}
\label{sup_sec:limitations}

There are some risks with using cross-modal pretraining on large-scale video datasets.
The distribution of identities and activities in the video dataset may not be representative of the global human population and the diversity in society. The social, gender, racial, and other biases in the dataset could be amplified during pretraining and knowledge distillation.
Also, the video dataset may include some private information, which could be vulnerable to dataset extraction attacks \cite{extractionattack}.
Moreover, our teacher model learns multi-modal grounding via contrastive learning between video and text tokens. However, each text token describes only certain parts of videos. The errors in multi-modal grounding would also be propagated to student models during knowledge distillation, hence we recommend careful use for real-world applications (similar to previous works in video understanding).

\end{document}